\documentclass{bmvc2k}
\usepackage{algorithm}
\usepackage{algorithmic}
\usepackage{esvect}
\usepackage{multirow}
\usepackage{pifont}

\title{Re-localization acceleration with Medoid Silhouette Clustering}

\addauthor{Hongyi Zhang}{cd20255@bristol.ac.uk}{1}
\addauthor{Walterio Mayol-Cuevas}{walterio.mayol-cuevas@bristol.ac.uk}{1,2}

\addinstitution{
 Visual Information Laboratory\\
 University of Bristol\\
 Bristol, UK
}
\addinstitution{
 Amazon, Seattle, USA
}

\runninghead{ZHANG ET AL.}{Re-localization acceleration for neural network}

\begin{document}

\maketitle

\begin{abstract}
Two crucial performance criteria for the deployment of visual localization are speed and accuracy. Current research on visual localization with neural networks is limited to examining methods for enhancing the accuracy of networks across various datasets. How to expedite the re-localization process within deep neural network architectures still needs further investigation. In this paper, we present a novel approach for accelerating visual re-localization in practice. A tree-like search strategy, built on the keyframes extracted by a visual clustering algorithm, is designed for matching acceleration. Our method has been validated on two tasks across three public datasets, allowing for 50 up to 90 percent time saving over the baseline while not reducing location accuracy.
\end{abstract}
\section{Introduction}
\label{sec:intro}
Visual place recognition plays a critical role in robotics and computer vision. Traditional Algorithms, including SIFT\cite{Lowe2004DistinctiveIF} and ORB\cite{Rublee2011ORBAE}, are utilised in SLAM systems to extract scene features. These algorithms are capable of scale and rotation invariant tolerance which allows for robust operation. Furthermore, learning algorithms such as bags-of-visual-words\cite{Philbin2007ObjectRW} and VLAD\cite{Arandjelovi2013AllAV} aggregate features into vectors for robust re-localization. Researchers are not only concerned with the development of reliable algorithms for accurate visual place recognition, but also the speed of re-localization. For instance, by restricting their Bayesian model, Mark and Paul utilised sparse matching between locations and visual words for the reduction of calculation in re-localization\cite{Cummins2011AppearanceonlySA}.\\
In recent years, deep convolutional neural networks (DCNNs) have emerged as the dominant paradigm in computer vision for pattern recognition, due to their exceptional efficacy in object detection and recognition. Furthermore, convolutional neural networks are employed by researchers to address the issue of visual place recognition. Several visual place recognition models(VPR models), such as NetVLAD\cite{Arandjelovi2015NetVLADCA}, DINOv2 SALAD\cite{Izquierdo2023OptimalTA}, Patch-NetVLAD\cite{Hausler2021PatchNetVLADMF}, MixVPR\cite{Alibey2023MixVPRFM}, and CosPlace\cite{Berton2022RethinkingVG}, are trained on large datasets including Mapillary\cite{Warburg2020MapillarySS}, Pittsburgh-30k\cite{Torii2013VisualPR}, or the Oxford RobotCar Dataset\cite{Maddern20171Y1}. Test results indicate that the accuracy of matching remains high, irrespective of variations in lighting, angle, or seasons. Nevertheless, the majority of research related to VPR models focuses on the enhancement of the model's accuracy. Methods for reducing the calculation required for re-localization with the VPR model are not yet under investigation.\\
The main contributions of our work are as follows: 1) A new method, compatible with the VPR model, aims at reducing the amount of computation while maintaining accuracy during the process of visual matching for re-localization. 2) Assessment of our approach using various benchmark datasets and relevant criteria. As well as additional relevant approaches are assessed to compare them with our proposed method.
\vspace{-1em}
\section{Related Work}
\label{sec:work}
\subsection{Visual place recognition neural network}
The issue of visual place recognition has frequently been categorised as a classification problem. The query image is initially compared with every image in the database. Prior to matching, the images are typically transformed into normalised vectors using the VPR model. The position of the query image is then estimated using the coordinate of the most visually similar image.\\
The objective of VPR model training is to increase the distance between vectors of images captured from different locations while decreasing the distance between that from the same location under different viewpoints over time. In recent years, a variety of research reports pertaining to the enhancement of the VPR model's accuracy have been published. For instance, Relja drew inspiration from the Vector of Locally Aggregated Descriptors representation and positioned the module responsible for relocating a cluster centre at the end of the CNN Network\cite{Arandjelovi2015NetVLADCA,Jgou2010AggregatingLD}. This modification significantly enhances the accuracy of the VPR model.
\vspace{-1em}
\subsection{Mapping and re-localization}
Researchers are mainly concerned with developing new VPR models to build robust features for re-localization. We note that the acceleration of re-localization, which enables practical deployment of performant visual algorithms is rarely investigated. In contrast, technologies related to fast mapping and re-localization are often developed and integrated into the SLAM system.\\
The FastSLAM system incorporated a novel concept known as {\em Landmarks} in order to reduce the computational burden associated with mapping and re-localization\cite{Montemerlo2002FastSLAMAF}. Their Victoria Park experiments show that the addition of an excessive number of inaccurate landmarks to the map lengthens the re-localization process. Sebastian invented a technique to eliminate landmarks that were supported by scant evidence gathered from the sensors, thereby expediting the operation of total system\cite{Thrun2004FastSLAMAE}.\\
The work of Contreras et.al. \cite{Contreras2015TrajectorydrivenPC} utilises splines and NURBS parameterization to fit the trajectory and construct the map via solely the features surrounding the {\em Control Point} in an attempt to reduce the size of the map created and fill gaps between features.\\ 
These works employ {\em Landmarks} or {\em Control Points} to determine which features should be utilised to represent a particular region for decreasing the map size and expediting the re-localization process. These concepts invited us to consider how to locate these {\em Landmarks} in a video-sampled image sequence and how to use that information to accelerate the re-localization process within the framework of the VPR model.
\vspace{-1em}
\section{Methodology}
\label{sec:method}
\begin{figure}
\centering
\includegraphics[width=10cm]{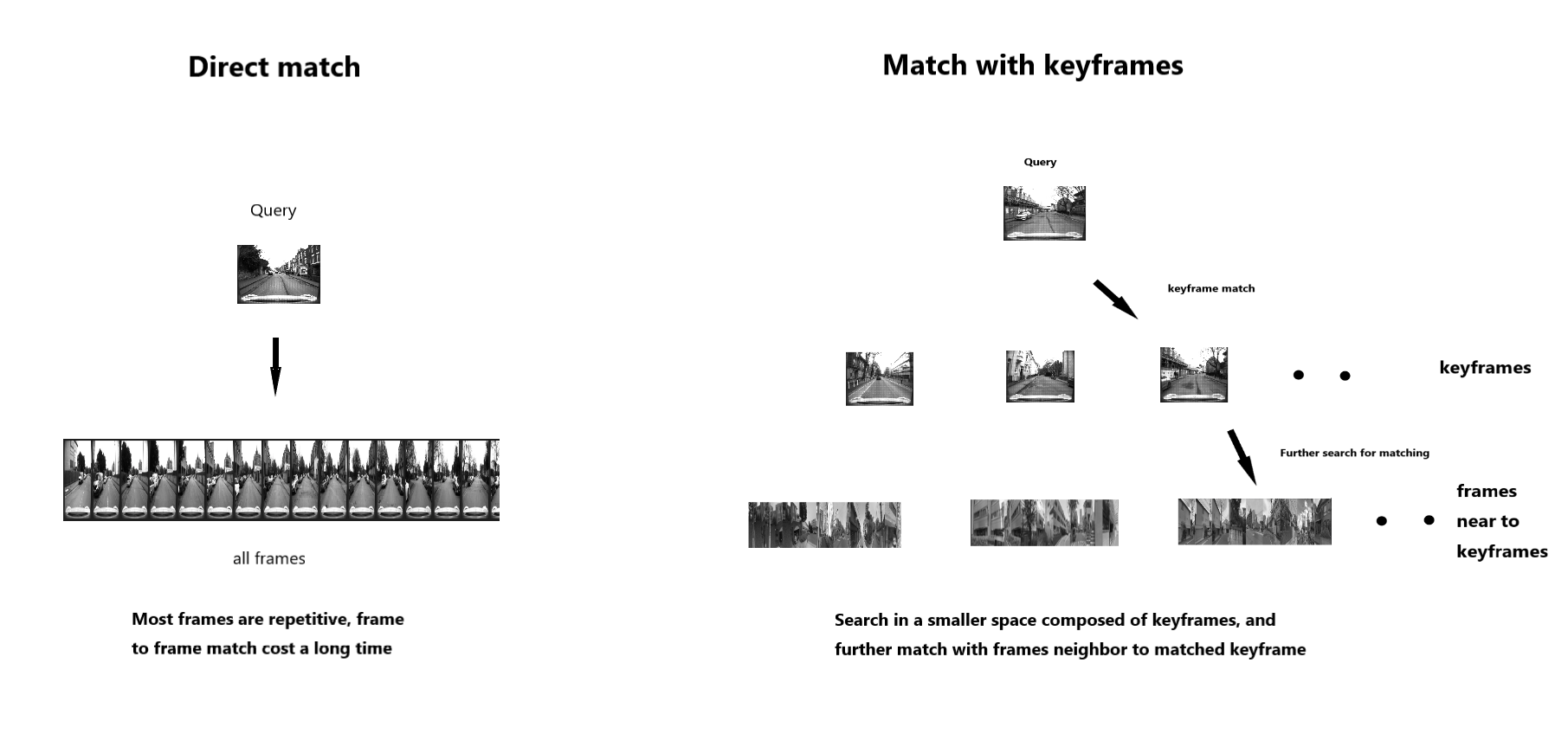}
\caption{Illustration of conventional re-localization and structured search via medioid keyframes}
\label{fig:illustrate}
\end{figure}
 In recent developments of model-based Visual Place Recognition, the matching process involves comparing the query image or query image sequence with every frame in the database, which largely lengthens the time of re-localization. We will explore how to reduce it within the VPR model framework and evaluate the approach in the experiment section.
 \vspace{-1em}
\subsection{Re-localization Acceleration}
The fundamental principle underlying the VPR model is to converge feature vectors originating from the same location while separating that place from different locations. This process converts the images sampled from video into a voronoi pattern, which consists of distinct clusters. Re-localization time can be significantly reduced by matching between the inquiry and {\em Keyframes} that are closest to the centre of these clusters at first. The optimal match can then be searched within the cluster or its neighbouring area. The procedure resembles the tree-like search illustrated in Figure \ref{fig:illustrate}. How to determine these clusters and the corresponding keyframes that represent them becomes the question. Re-localization with keyframes consists of two phases: keyframe extraction and further matching.
\vspace{-1em}
\subsection{Methods of Keyframes Extraction}
\subsubsection{Faster Medoid
Silhouette Clustering}
In this study, keyframes (medoids) are extracted from the image sequence using Faster Medoid Silhouette Clustering\cite{Schubert2020FastAE,Lenssen2023MedoidSC}. Alternative clustering techniques can also be deployed here to extract keyframes. A set of Medoids, directly coming from the dataset, will be acquired through a series of iterations of evaluation and swapping. This procedure comprises three steps in brief: (1) Initialize medoids (2) Repeatedly evaluate the Average Medoid Silhouette(AMS) of eager swapping between medoids and non-medoids, and excute the swap that resulted in an improved AMS (3) Terminate the algorithm when no swap exists to increase AMS.\\
The outcomes of the Faster Medoid Silhouette Clustering algorithm vary based on the initialization of the medoids. A common approach to the initialization involves selecting medoids at random from the entire dataset. Additionally we compare it with a naive strategy of uniform sampling at a fixed rate.\\
The score $s_{i}$ of non-medoid $i$ is measured by its distance $a_{i}$ to its closest medoid, its distance $b_{i}$ to its second closest medoid, and $b_{i} >= a_{i}$ as follows \\
\begin{equation}
\begin{split}
s_{i}(X,d,M) &= \frac{b_{i} - a_{i}}{max(a_{i},b_{i})} = \frac{b_{i} - a_{i}}{b_{i}} = 1 - \frac{a_{i}}{b_{i}} \\
a_{i} &= d(m_{l_{1}}, x_{i}) \\
b_{i} &= d(m_{l_{2}}, x_{i})
\end{split}
\end{equation}
where $X$ is the set of samples not belonging to the medoids, $d$ is the function to calculate the distance between non-medoids and medoids. $M$ is the set of medoids. $m_{l_{1}}$ is the closest medoid to non-medoid $i$, and $m_{l_{2}}$ is the second closest medoid to $i$. \\
Average Medoid Silhouette of a set of medoids is averaged over the scores of these non-medoids as follows\cite{Rousseeuw1987SilhouettesAG}.
\begin{equation}
\tilde{S}(X,d,M) = \frac{1}{n} \sum_{i=1}^{n} s_{i}(X,d,M)
\end{equation}
The features generated by the VPR model\cite{Izquierdo2023OptimalTA} we use are normalized vectors, and cosine similarity is often used to measure the distance between two features. Here, we define the distance function for two features as $d(\vv{a},\vv{b}) = 1 - cos(\vv{a},\vv{b})$, and pseudocode of keyframes extraction is as follows.
\begin{algorithm} 
        \renewcommand{\algorithmicensure}{\textbf{Output:}}
	\caption{Keyframes extraction with Faster Medoid
Silhouette Clustering} 
	\label{key} 
	\begin{algorithmic}[1]
		\REQUIRE VPR model $M$, Image Sequence sampled from video $Seq$
		\STATE Extract features with $M$: $fseq$ = $M$($Seq$)
            \STATE Initialize $Keyframes$; Split $fseq$ into medoids $fkeys$ and non-medoids $fsamples$
            \STATE Define: $d(\vv{a},\vv{b})$ = 1 - $cos(\vv{a},\vv{b})$
            \STATE $Keyframes$ = $fasterMSC(fsamples, d, fkeys)$
            \ENSURE $Keyframes$
	\end{algorithmic} 
\end{algorithm}
\vspace{-2em}
\subsubsection{Other Basic Methods}
The performance of three alternative methods for selecting keyframes is also evaluated in the experiment section.\\
$1)$ Similarity change: In a continuous image sequence, if the cosine similarity between the current frame and the last selected key frame falls below a specified threshold, the current frame will be designated as the new key frame.\\
$2)$ Distance change on trajectory: when the current frame's geographical distance from the previous keyframe as measured with GPS exceeds a specified threshold, it is designated as the new keyframe.\\
$3)$ Fixed frame rate: keyframes are sampled from the image sequence(database) at a fixed frame rate, which is determined by the number of keyframes.
\vspace{-1em}
\section{Experiments and Evaluation}
In order to assess the accuracy of deep neural network-based visual place recognition with keyframes, our approach was evaluated across various datasets and compared with the baseline of VPR techniques. In this work, our target is not to evaluate the network itself. Alternatively, we investigate how the accuracy changes when keyframes are implemented to accelerate the re-localization in comparison with the baseline.
\vspace{-1em}
\subsection{Basic techniques}
\begin{figure}
\centering
\includegraphics[width=12cm]{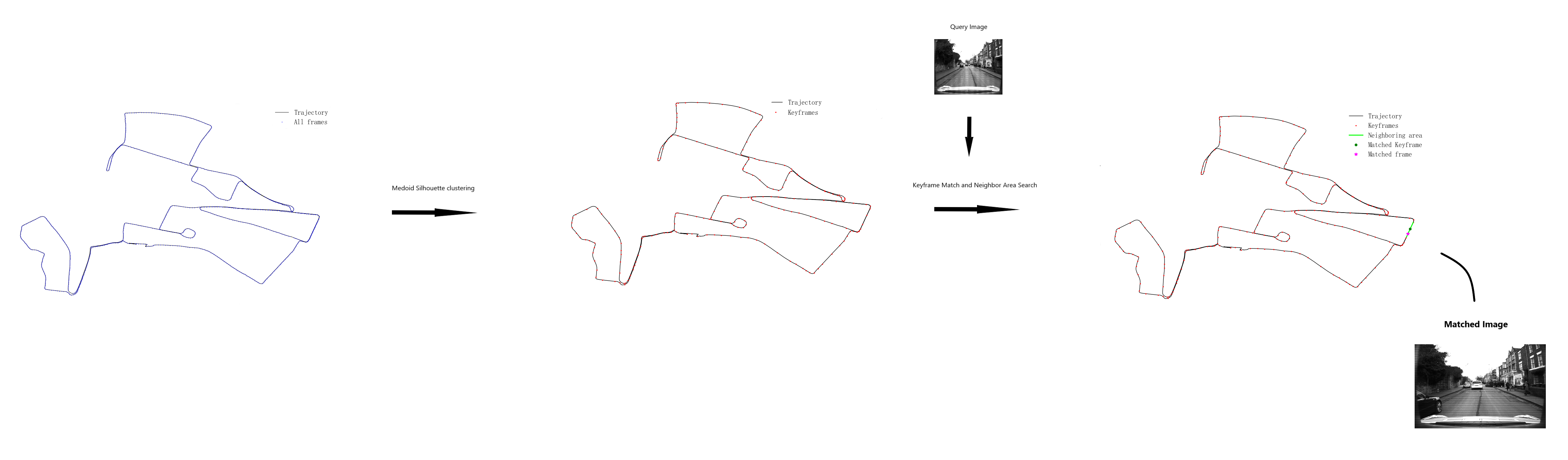}
\caption{Visualization of re-localization with Keyframes: keyframes extraction, Match query with keyframes, further neighboring area for accurate position.}
\label{fig:relocate_key}
\end{figure}
The features of images in our study are generated by the VIT model $DINOv2$ $SALAD$, which ranks top on the $Mapillary$ dataset\cite{Izquierdo2023OptimalTA}. The baseline for comparison is the im2im and seq2seq (sequence length is 3) tasks described in $Mapillary$ without deploying our methods\cite{Warburg2020MapillarySS}.\\
Three additional approaches, keyframe selection by distance change, similarity change, and fixed frame rate, are assessed in comparison with the clustering methodology for keyframes extraction.\\
 As mentioned in \ref{sec:method}, diverse approaches to initializing medoids may yield varying outcomes. In this section, two methods for initialization are introduced and evaluated here. 1) The initial keyframes are chosen at random from the image sequence, and the clustering approach is applied to find the best keyframes. The keyframes with the highest Average Medoid Silhouette are chosen for use after this procedure is repeated ten times. 2) Depending on the quantity of keyframes required, we sample the initial keyframes at a constant frame rate.\\
Utilizing our methodologies, the im2im and seq2seq tasks are illustrated below. In the im2im task, the query image is initially compared with the keyframes, and subsequently with the images in the adjacent region that contain the keyframe with the highest similarity. In the task of seq2seq, each keyframe is compared with every image in the query sequence and given a summation score. The further seq2seq matching is performed in adjacent regions of the highest-scoring keyframe. As shown in the right image of Figure \ref{fig:relocate_key}, the adjacent region in our study is the image sequence, which begins from the prior keyframe and terminates at the succeeding keyframe.
\vspace{-1em}
\subsection{Datasets}
Three benchmark datasets have been used to evaluate our methodology: the Nordland dataset\cite{olid2018single}, the Gardens Point Walking dataset\cite{Chen2014MultiscaleBP} and the Oxford Radar RobotCar Dataset\cite{RadarRobotCarDatasetICRA2020}.\\
The Nordland dataset contains single-view long image sequences captured from Nordland railway in four seasons\cite{olid2018single}. The experiment utilizes image sequences labelled 1395-14232 across four seasons. Keyframes are extracted from the spring dataset for the construction of search tree. Image or image sequences, sampled from dataset of three other seasons, are utilized as the image query or sequence query. Gardens Point Walking is a image sequence dataset containing 200 image pairs along the same route under different viewpoints as captured at a university\cite{Chen2014MultiscaleBP}. The search tree is constructed via keyframes from the right-hand dataset, while the query utilizes images or image sequences from the left-hand dataset.\\
Oxford Radar RobotCar Dataset is comprised of images collected by various cameras on the vehicle, radar information of the area, and GPS location data in the form of latitude and longitude for each frame along a 9-kilometer route through the heart of Oxford under varying weather conditions and different days\cite{RadarRobotCarDatasetICRA2020, Maddern20171Y1}. GPS data of the frames and sequences of images captured by the center camera are the only components utilized. 1063 images are selected from a single image sequence containing over 30,000 images, with the choosing determined by the GPS position. Current frame is added to the database when the Manhattan distance between the GPS locations of the current frame and the last chosen frame is greater than 0.0001 degree. With the GPS coordinates and chronological ordering of the images, we find the most optimal matches of the selected images in the data captured from a different time period.
\vspace{-1em}
\subsection{Results}
The evaluation of two tasks performed with and without the implementation of keyframes consists of two components: query accuracy and time expenditure.
\vspace{-1em}
\subsubsection{Computational time cost}
\begin{figure}
\centering
\includegraphics[width=8cm]{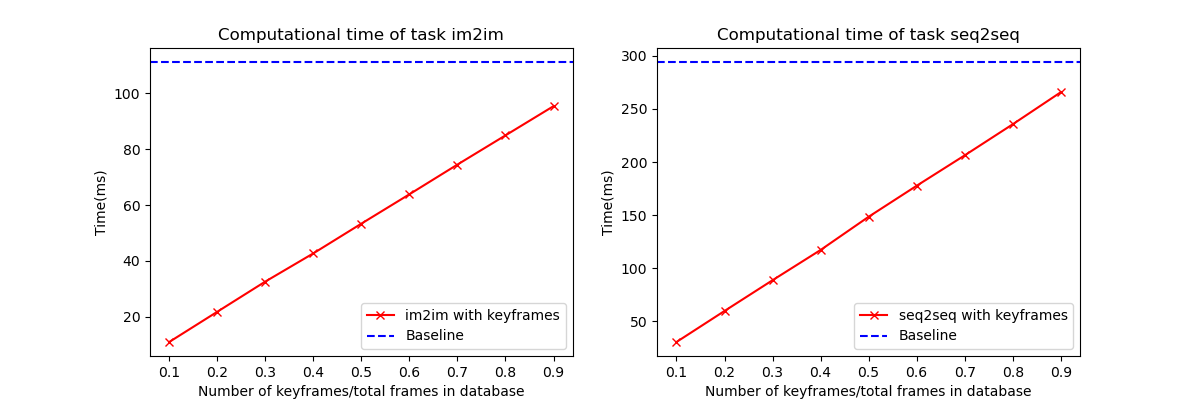}
\caption{Computational time of im2im and seq2seq task}
\label{fig:ctime}
\end{figure}
\begin{table}
\begin{center}
\scalebox{0.7}{
\begin{tabular}{ |c||c|c|c|c|c|c|  }
 \hline
 \multicolumn{7}{|c|}{Computational time cost per query} \\
 \hline
Task & ratio 0.1 & ratio 0.2 & ratio 0.3 & ratio 0.4 & ratio 0.5 & Baseline\\
 \hline
 Im2Im& 10.96ms & 21.78ms & 32.59ms & 42.69ms & 53.33ms & 111.2ms\\
 \hline
 Seq2Seq& 30.68ms & 59.87ms & 88.80ms & 117.25ms & 148.48ms & 294.07ms\\
 \hline
\end{tabular}}
\end{center}
\caption{Time cost of query with and without keyframes on the tasks of im2im and seq2seq}
\label{table:timecost}
\end{table}
On the Nordland dataset, time evaluations of the tasks im2im and seq2seq are performed with and without keyframes. The features extracted from a continuous image sequence comprising 12836 spring images are utilized as the database. 12836 times of query, from the dataset of summer, autumn and winter, have been done on the tasks mentioned above separately. For the seq2seq task, the length of sequence is 3 in our work. The average matching time for the two tasks was determined using an Intel(R) Core(TM) i7-6700K processor operating at 4.00GHz. In Figure \ref{fig:ctime}, the time required to complete two tasks evaluated on the Nordland dataset is depicted. The figure's x-axis corresponds to the ratio of keyframes among all frames in the database, while the y-axis denotes the computational time in milliseconds. As the ratio rises from 0.1 to 0.9, the time required to complete an im2im match with our keyframe module increases almost linearly indicated by the red line on the left of Figure \ref{fig:ctime}. The time of seq2seq task with our module is represented by the red line on the right of Figure \ref{fig:ctime}. This line also displays a linear growth trajectory, increasing from 0.1 ratio to 0.9 ratio. Two blue lines in Figure \ref{fig:ctime} indicate the baselines of two tasks conducted without deploying our module. Both tasks utilizing our module exhibit a significant reduction in query time when compared to the baseline.\\
The comprehensive computational time costs are detailed in Table \ref{table:timecost}. The ratio from the second to sixth columns in the table is calculated by dividing the number of keyframes by the total number of frames in the database. The time required to complete the tasks without implementing our method is denoted in the table as the baselines.\\
Figure \ref{fig:ctime} and Table \ref{table:timecost} illustrate the time-saving potential of our module for two tasks. Furthermore, we conduct additional analysis to assess the accuracy of image querying with keyframes for these two tasks in the following sections.
\vspace{-1em}
\subsubsection{Accuracy Evaluation}
\begin{figure}
\centering
\includegraphics[width=8cm]{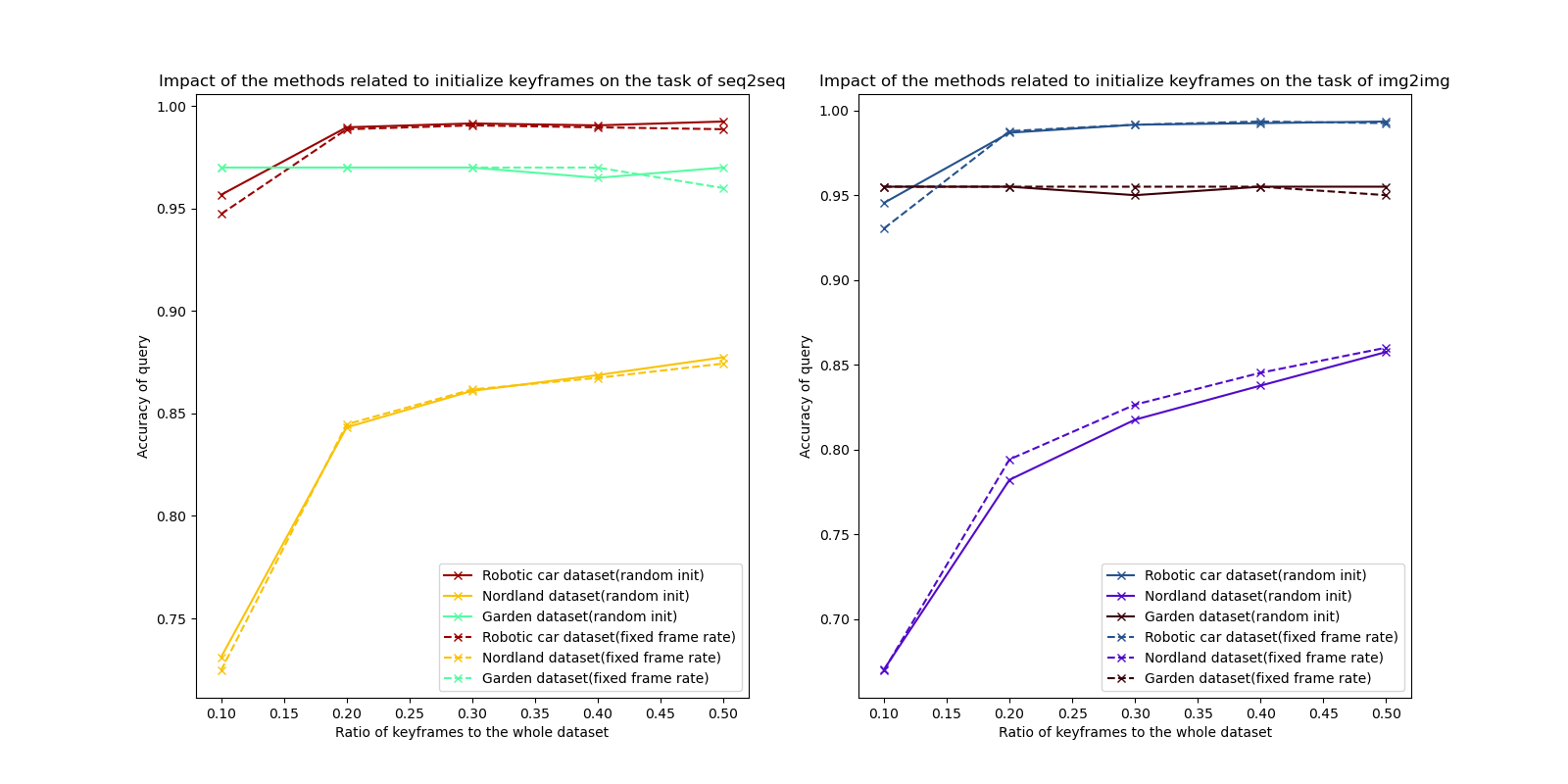}
\caption{Accuracy of two tasks with different methods to initializing the keyframes}
\label{fig:init}
\end{figure}

\begin{table}
\begin{center}
\scalebox{0.7}{
\begin{tabular}{ |c|c||c|c|c|c|c|c|  }
 \hline
Dataset & Task & ratio 0.1 & ratio 0.2 & ratio 0.3 & ratio 0.4 & ratio 0.5 & Baseline\\
 \hline
 \multirow{2}*{robot car} & Im2Im & 0.934 & 0.988 & 0.992 & 0.987 & 0.992 & 0.995\\
 \cline{2-8}
 ~ & Seq2Seq & 0.943 & 0.991 & 0.992 & 0.984 & 0.989 & 0.992\\
 \hline
  \multirow{2}*{Garden} & Im2Im & 0.955 & 0.955 & 0.955 & 0.955 & 0.950 & 0.955\\
 \cline{2-8}
 ~ & Seq2Seq & 0.970 & 0.970 & 0.970 & 0.960 & 0.960 & 0.970\\
 \hline
  \multirow{2}*{Nordland} & Im2Im & 0.664 & 0.784 & 0.817 & 0.831 & 0.853 & 0.903\\
 \cline{2-8}
 ~ & Seq2Seq & 0.724 & 0.845 & 0.860 & 0.865 & 0.874 & 0.935\\
 \hline
\end{tabular}
}
\end{center}
\caption{Accuracy of image query on three different datasets with keyframes}
\label{table:acc}
\end{table}
\begin{figure}
\centering
\includegraphics[width=8cm]{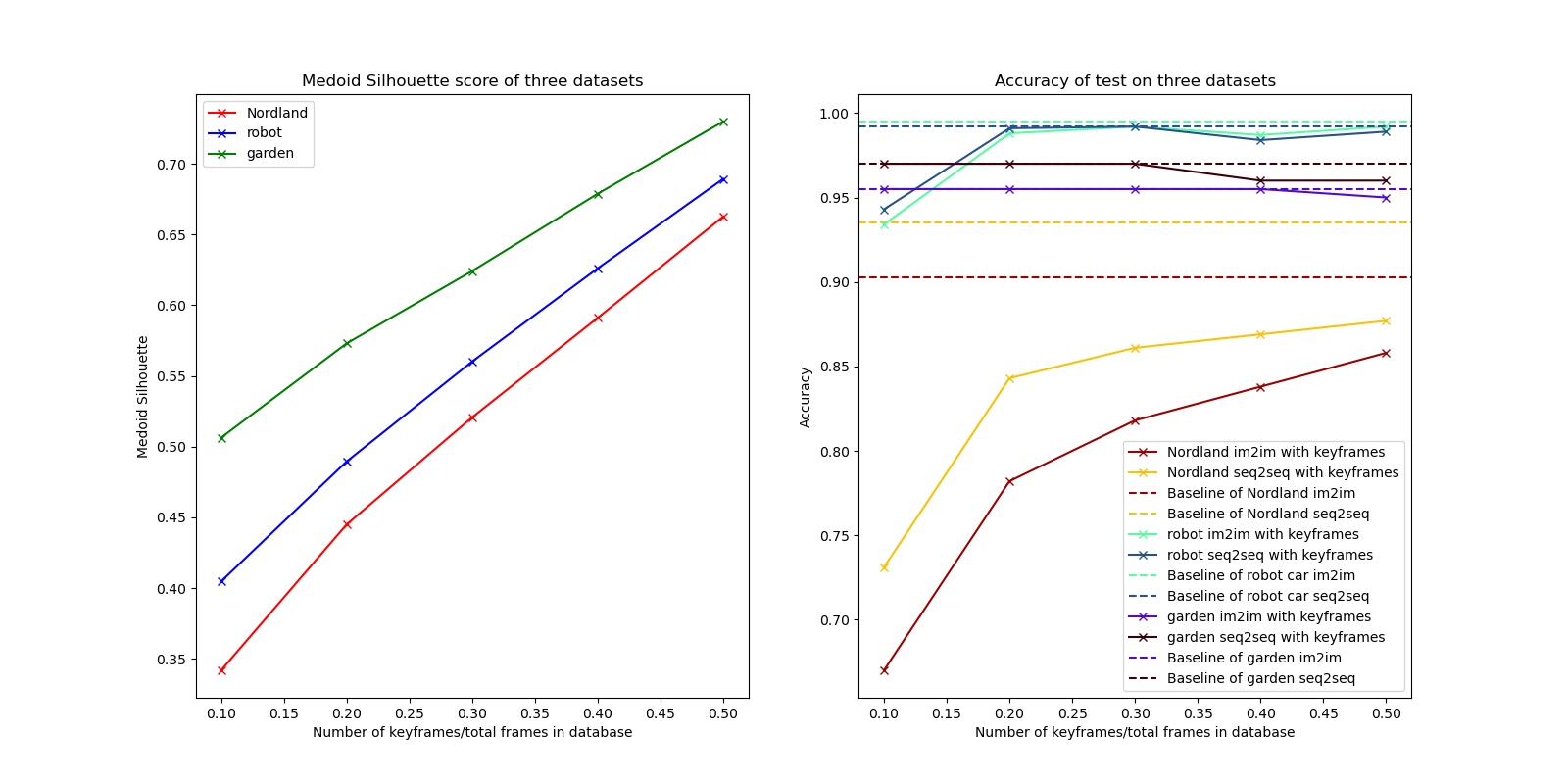}
\caption{Medoid Silhouette and Accuracy of tests on three datasets}
\label{fig:medoid}
\end{figure}
Based on three benchmark datasets, we assessed the accuracy of our methodologies implemented on two tasks. For the previously mentioned Nordland dataset, the spring image sequence serves as the database from which the keyframes are generated. The accuracy of the test is calculated by averaging the results of the query across the Nordland data of three other seasons. Each image in the Gardens Point Walking and Norland Dataset is paired with a corresponding matching image, enabling querying and accuracy calculation. Each frame in the Oxford Radar car Dataset is associated with GPS data, which enables the computation of accuracy in relation to distance. We establish a +/-2 frame tolerance for true matches in the Nordland and Gardens Point Walking dataset test, and a +/-0.0002 degree tolerance in GPS location for correct matches on the test of Oxford Radar Robot Car dataset.\\
The proportion of keyframes to frames in the database is denoted in Table \ref{table:acc} as the ratio. As shown in Table \ref{table:acc}, querying with keyframes performs well on the Garden and Robot Radar Car datasets, achieving high accuracy of matching even though the ratio is very low. The keyframes' performance on the Nordland dataset is inferior to that on two other datasets. The problem relating to the performance of keyframes on the Nordland dataset is analyzed by the Average Medoid Silhouette, as shown in Figure \ref{fig:medoid}. Low Average Medoid Silhouette represents the quality of the clustering result is poor, leading to performance degradation of our method. The Average Medoid Silhouette of the Nordland dataset exhibits a lower value compared with that of the Garden Point Walking and Oxford Robotic Radar Car datasets, as evidenced by a gap between its accuracy and the baseline (right image in Figure \ref{fig:medoid}). The discrepancy between the accuracy of our method and the baseline is close to zero when the keyframe-to-database ratio exceeds 0.2 on the robotic car dataset. Among all datasets, the Average Medoid silhouette of the garden dataset is the highest. The test results of the garden dataset indicate that there is no discrepancy between the accuracy of our method and the baseline, even when the keyframe ratio is 0.1. Overall, the Average Medoid silhouette can be utilized to assess the score of keyframes. Whether or not to utilize keyframes for re-localization can be determined by analyzing their quality.\\
In the evaluation described above, initial keyframes are selected from the database of the image sequence at a fixed frame rate, which is determined by the keyframe-to-database ratio. An alternative approach to initializing keyframes involves conducting a uniform random sample from the database several times. The keyframes with highest Average Medoid Silhouette after clustering are reserved. The outcomes of these two methods are illustrated in Figure \ref{fig:init}. The performance disparity between the method initialized at a fixed frame rate (dash line) and that initialized at random (solid line) is not much for both tasks. Overall, the outcomes of these tasks are not significantly impacted by the aforementioned initialization methodologies.
\begin{table}[t]
\begin{center}
\scalebox{0.7}{
\begin{tabular}{ |c||c|c|c|  }
 \hline
Methods & Trajectory free & Specified number & Quality criterion \\
 \hline
 Medoid Silhouette & \ding{52} & \ding{52} & \ding{52} \\
 \hline
 Fixed frame rate & \ding{52} & \ding{52} & \ding{56} \\
 \hline
 Cosine Similarity & \ding{52} & \ding{52} & \ding{56} \\
 \hline
 Distance & \ding{56} & \ding{52} & \ding{56} \\
 \hline
\end{tabular}
}
\end{center}
\caption{Comparison between different keyframes selection strageties}
\label{table:star}
\end{table}
\subsubsection{Evaluation of methods on selecting keyframes}
\begin{figure}
\centering
\includegraphics[width=8cm]{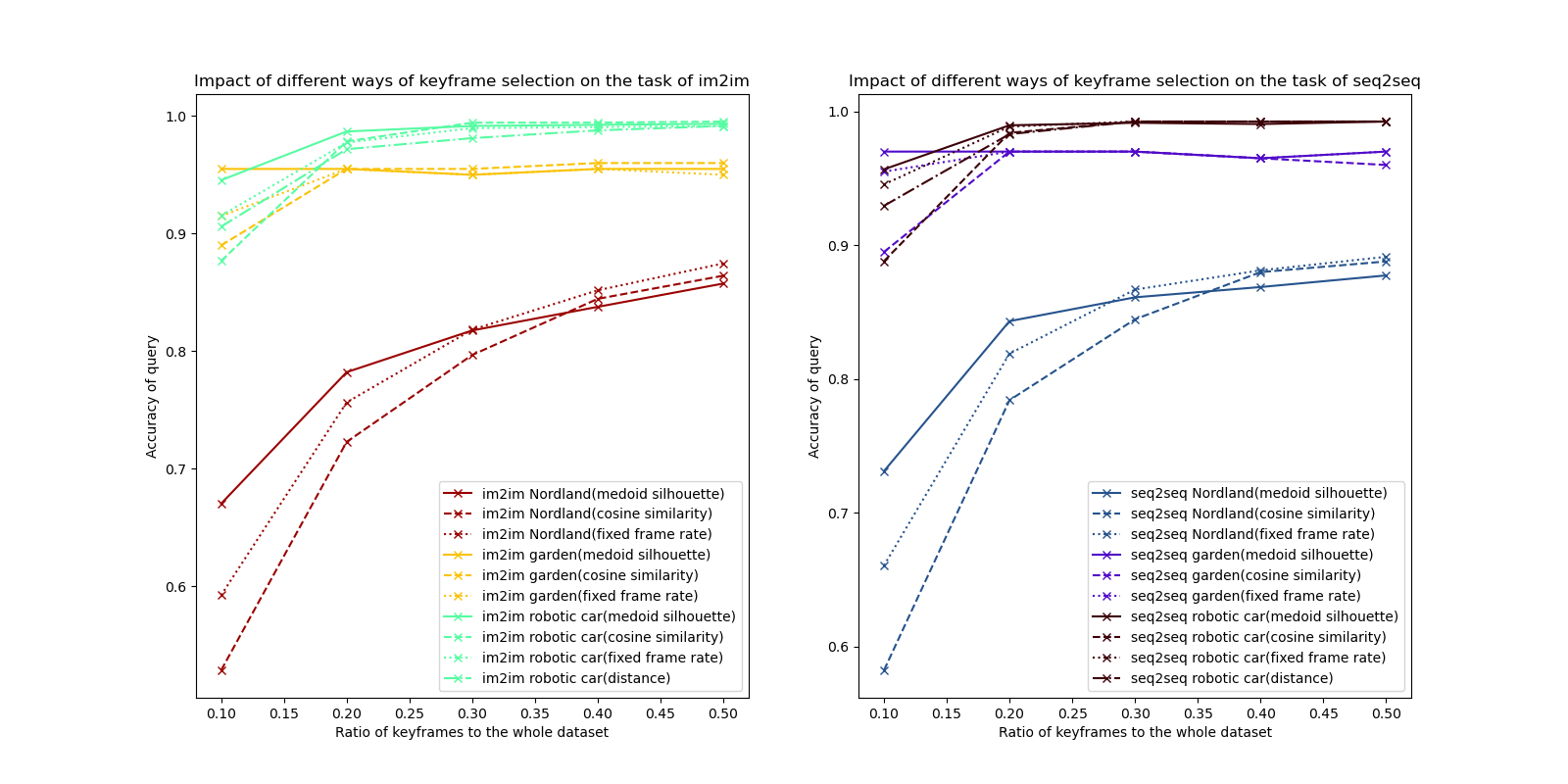}
\caption{Accuracy of two tasks with different methods to extract keyframes}
\label{fig:keyframe}
\end{figure}

In addition to the clustering method for selecting keyframes, three additional methods mentioned in methodology for extracting keyframes are also assessed from various perspectives. Trajectory-free in Table \ref{table:star} indicates that additional trajectory information is not needed. A specified number indicates that any number of keyframes can be selected. The quality criterion serves as an indicator for assessing the reliability of keyframe re-localization.\\ 
As shown in Table \ref{table:star}, additional trajectory information is required for keyframe selection with distance. Each of these approaches has the ability to choose a certain number of keyframes. However, the cosine similarity method requires a high precision threshold in order to closely match the desired number, and there is no guarantee that it will actually reach that number. Unlike Faster Medoid Silhouette clustering, these three algorithms lack a criterion for evaluating the quality of the keyframes. The reliability of re-localization with these keyframes is uncertain without the quality criterion.\\
The GPS information is only available in the Oxford radar car dataset, allowing us to extract the keyframes based on distance. Other datasets lack the required data for extracting keyframes based on distance. Figure \ref{fig:keyframe} demonstrates that the faster medoid silhouette clustering method surpasses three other methods in terms of accuracy when the ratio of keyframes is low for both tasks. With the exception of the Nordland dataset, the accuracy of four approaches converge to the same value as the ratio increases. In the Nordland dataset, other methods outperform the faster medoid silhouette clustering slightly on two tasks at high ratios. In table \ref{table:RUA}, area under the accuracy curve also demonstrates that the clustering method performs better than three other methods in the aspect of re-localization accuracy. 
\begin{table}[t]
\begin{center}
\scalebox{0.7}{
\begin{tabular}{ |c|c||c|c|c|c| }
 \hline
Dataset & Task & Medoid & Similarity & Fixed frame& Distance\\
 \hline
 \multirow{2}*{Robot car} & Im2Im & \pmb{0.394} & 0.390 & 0.391 & 0.388\\
 \cline{2-6}
 ~ & Seq2Seq & \pmb{0.395} & 0.391 & 0.394 & 0.392 \\
 \hline
  \multirow{2}*{Garden} & Im2Im & \pmb{0.382} & 0.379 & 0.379 & /\\
 \cline{2-6}
 ~ & Seq2Seq & \pmb{0.388} & 0.383 & 0.386 & /\\
 \hline
  \multirow{2}*{Nordland} & Im2Im & \pmb{0.320} & 0.306 & 0.316 & /\\
 \cline{2-6}
 ~ & Seq2Seq & \pmb{0.338} & 0.324 & 0.334 & /\\
 \hline
\end{tabular}
}
\end{center}
\caption{Area under 
accuracy curve of tasks with different methods for keyframe selection}
\label{table:RUA}
\end{table}
\vspace{-1em}
\section{Conclusion and Future work}
\vspace{-0.5em}
In this paper, we present a novel approach to enhance the speed of model-based re-localization and assess the effectiveness of this approach as well as other existing methods across various datasets. The introduction of keyframes in the tasks significantly speeds up the process of matching. In addition, our clustering method for selecting keyframes surpasses three other methods in the aspect of accuracy. Furthermore, it offers a criterion to assess the quality of these keyframes, enabling us to determine the reliability of re-localization with these keyframes. While our method helps expedite the re-localization process, there is still a gap in accuracy compared to the non-compressed baseline method, particularly when used on the Nordland dataset. Finally, our approach is aimed at improving performance specifically on embedded systems an area we believe merits further development to bring applied computer vision to fruition.
\bibliography{egbib}
\end{document}